# Linking Symptom Inventories using Semantic Textual Similarity

Eamonn Kennedy[1-3], Shashank Vadlamani[1], Hannah M Lindsey[1,2], Kelly S Peterson[3,4], Kristen Dams O'Connor[5,6], Kenton Murray[7], Ronak Agarwal[1], Houshang H Amiri[8,9], Raeda K Andersen[10,11], Talin Babikian[12,13], David A Baron[14], Erin D Bigler[1,15,16], Karen Caeyenberghs[17], Lisa Delano-Wood[18,19], Seth G Disner[20,21], Ekaterina Dobryakova[22,23], Blessen C Eapen[24,25], Rachel M Edelstein[26], Carrie Esopenko[5], Helen M Genova[27], Elbert Geuze[28], Naomi J Goodrich-Hunsaker[1], Jordan Grafman[29,30], Asta K Håberg[31,32], Cooper B Hodges[33], Kristen R Hoskinson[34,35], Elizabeth S Hovenden[1], Andrei Irimia[36-38], Neda Jahanshad[39], Ruchira M Jha[40], Finian Keleher[1], Kimbra Kenney[41,42], Inga K Koerte[43,44], Spencer W Liebel[1], Abigail Livny,[45,46] Marianne Løvstad[47,48], Sarah L Martindale[49,50], Jeffrey E Max[51], Andrew R Mayer[52], Timothy B Meier[53], Deleene S Menefee[54,55], Abdalla Z Mohamed[56], Stefania Mondello[57], Martin M Monti[58,59], Rajendra A Morey[60,61], Virginia Newcombe[62], Mary R Newsome[1,2], Alexander Olsen[63-65], Nicholas J Pastorek[54,66], Mary Jo Pugh[67,68], Adeel Razi[69-71], Jacob E Resch[72], Jared A Rowland[49,73,74], Kelly Russell[75,76], Nicholas P Ryan[17,77], Randall S Scheibel[54,66], Adam T Schmidt[78,79], Gershon Spitz[80,81], Jaclyn A Stephens[82,83], Assaf Tal[84], Leah D Talbert[33], Maria Carmela Tartaglia[85,86], Brian A Taylor[87], Sophia I Thomopoulos[39], Maya Troyanskaya[54,66], Eve M Valera[88,89], Harm Jan van der Horn[52], John D Van Horn[90], Ragini Verma[91,92], Benjamin SC Wade[93], Willian SC Walker[94,95], Ashley L Ware[96,97], J Kent Werner Jr[41], Keith Owen Yeates[98], Ross D Zafonte[99,100], Michael M Zeineh[101], Brandon Zielinski[1,102,103], Paul M Thompson[39,104], Frank G Hillary[105,106], David F Tate[1,2], Elisabeth A Wilde[1,2]*, Emily L Dennis[1,2]*

1. Department of Neurology, University of Utah School of Medicine, Salt Lake City, UT.
2. George E. Wahlen Veterans Affairs Medical Center, Salt Lake City, UT.
3. Division of Epidemiology, University of Utah, Salt Lake City, UT.
4. Veterans Health Administration (VHA) Office of Analytics and Performance Integration (VHA).
5. Department of Rehabilitation and Human Performance, Icahn School of Medicine at Mount Sinai, New York, NY.
6. Department of Neurology, Icahn School of Medicine at Mount Sinai, New York, NY.
7. Human Language Technologies Center of Excellence, Johns Hopkins University, Baltimore, MD.
8. Neuroscience Research Center, Institute of Neuropharmacology, Kerman University of Medical Sciences, Kerman, Iran.
9. Department of Radiology and Nuclear Medicine, Vrije Universiteit Amsterdam, Amsterdam UMC Location VUmc, Amsterdam, The Netherlands.
10. Crawford Research Institute, Shepherd Center, Atlanta, GA.
11. Department of Sociology, Georgia State University, Atlanta, GA.
12. Department of Psychiatry and Biobehavioral Sciences, Semel Institute for Neuroscience and Human Behavior, UCLA, Los Angeles, CA.
13. UCLA BrainSPORT Program, Los Angeles, CA.
14. Center for Behavioral Health and Sport, Dept. of Psychiatry, Western University of Health Sciences, Lebanon, OR and Pomona, CA.
15. Department of Psychology and Neuroscience Center, Brigham Young University, Provo, UT.
16. Department of Psychiatry, University of Utah, Salt Lake City, UT.
17. Cognitive Neuroscience Unit, School of Psychology, Deakin University, Geelong, Australia.
18. VA San Diego Healthcare System; Center of Stress and Mental Health, and Department of Psychiatry, UC San Diego School of Medicine, San Diego, CA.
19. Department of Psychiatry, UC San Diego School of Medicine, San Diego, CA.
20. Minneapolis VA Health Care System, Minneapolis, MN.
21. Department of Psychiatry, University of Minnesota, Minneapolis, MN.
22. Center for Traumatic Brain Injury, Kessler Foundation, East Hanover, NJ.
23. Rutgers New Jersey Medical School, Newark, NJ.
24. VA Greater Los Angeles Health Care System, Los Angeles, CA.
25. Division of Physical Medicine and Rehabilitation, David Geffen School of Medicine at UCLA, Los Angeles, CA.
26. Department of Psychology, University of Virginia, Charlottesville, VA.
27. Center for Autism Research, Kessler Foundation, East Hanover, NJ.
28. Brain Research and Innovation Centre, Ministry of Defence, Utrecht, The Netherlands.

29. Shirley Ryan Ability Lab, Chicago, IL.
30. Northwestern University, Chicago, IL.
31. Department of Neuromedicine and Movement Science, Faculty of Medicine and Health Sciences, Norwegian University of Science and Technology (NTNU), Trondheim, Norway.
32. Department for Physical Health and Aging, Norwegian Institute of Public Health, Oslo, Norway.
33. Department of Psychology, Brigham Young University, Provo, UT.
34. Center for Biobehavioral Health, Abigail Wexner Research Institute at Nationwide Children's Hospital, Columbus, OH.
35. Department of Pediatrics, The Ohio State University College of Medicine, Columbus, OH.
36. Ethel Percy Andrus Gerontology Center, Leonard Davis School of Gerontology, University of Southern California, Los Angeles, CA.
37. Alfred E. Mann Department of Biomedical Engineering, Viterbi School of Engineering, University of Southern California, Los Angeles, CA.
38. Department of Quantitative and Computational Biology, Dornsife College of Arts and Sciences, University of Southern California, Los Angeles, CA.
39. Imaging Genetics Center, Stevens Neuroimaging & Informatics Institute, Keck School of Medicine of USC, Marina del Rey, CA.
40. Departments of Neurology, Translational Neuroscience, Barrow Neurological Institute, Phoenix, AZ.
41. Department of Neurology, Uniformed Services University of the Health Sciences, Bethesda, MD.
42. National Intrepid Center of Excellence, Walter Reed National Military Medical Center, Bethesda, MD.
43. Psychiatry Neuroimaging Laboratory, Brigham & Women's Hospital, Boston, MA.
44. Department of Child and Adolescent Psychiatry, University Hospital, Ludwig-Maximilians-Universität, Munich, Germany.
45. Division of Diagnostic Imaging, Sheba Medical Center, Tel-Aviv, Israel.
46. Faculty of Medicine, Tel-Aviv University, Tel-Aviv, Israel.
47. Sunnaas Rehabilitation Centre, Nesodden, Norway.
48. University of Oslo, Oslo, Norway.
49. W. G. (Bill) Hefner VA Healthcare System, Salisbury, NC.
50. Department of Physiology & Pharmacology, Wake Forest School of Medicine, Winston-Salem, NC.
51. Department of Psychiatry, UC San Diego, La Jolla CA.
52. The Mind Research Network, Albuquerque, NM.
53. Department of Neurosurgery, Medical College of Wisconsin, Milwaukee, WI.
54. Michael E. DeBakey Veterans Affairs Medical Center, Houston, TX.
55. The Menning Department of Psychiatry and Behavioral Sciences, Baylor College of Medicine, Houston, TX.
56. Thompson Institute, University of the Sunshine Coast, Birtinya, Australia.
57. Department of Biomedical and Dental Sciences and Morphofunctional Imaging, University of Messina, Messina, Italy.
58. Department of Psychology, University of California Los Angeles, Los Angeles, CA.
59. Brain Injury Research Center (BIRC), Department of Neurosurgery, University of California Los Angeles, Los Angeles, CA.
60. Brain Imaging and Analysis Center, Duke University, Durham, NC.
61. Department of Psychiatry and Behavioral Sciences, Duke University, Durham, NC.
62. Department of Medicine, University of Cambridge, UK.
63. Department of Psychology, NTNU, Trondheim, Norway.
64. Clinic of Rehabilitation, St. Olavs Hospital, Trondheim University Hospital, Trondheim, Norway.
65. NorHEAD - Norwegian Centre for Headache Research, NTNU, Trondheim, Norway.
66. H. Ben Taub Department of Physical Medicine and Rehabilitation, Baylor College of Medicine, Houston, TX.
67. Information Decision-Enhancement and Analytic Sciences Center, VA Salt Lake City, Salt Lake City, UT.
68. Department of Medicine, University of Utah School of Medicine, Salt Lake City, UT.
69. Turner Institute for Brain and Mental Health, Monash University, Melbourne, Australia.
70. Wellcome Centre for Human Neuroimaging, University College London, London, United Kingdom.
71. CIFAR Azrieli Global Scholars Program, CIFAR, Toronto, Canada.
72. Department of Kinesiology, University of Virginia, Charlottesville, VA.
73. Department of Neurobiology and Anatomy, Wake Forest School of Medicine, Winston-Salem, NC.
74. Mid-Atlantic Mental Illness Research and Education Center, Durham, NC.
75. Department of Pediatrics and Child Health, University of Manitoba, Winnipeg, Canada.
76. Children's Hospital Research Institute of Manitoba, Winnipeg, Canada.
77. Brain and Mind Research, Murdoch Children's Research Institute, Parkville, Australia.
78. Texas Tech University Department of Psychological Sciences, Lubbock, TX.
79. Center of Excellence For Translational Neuroscience and Therapeutics, Lubbock, TX.
80. Monash-Epworth Rehabilitation Research Centre, School of Psychological Sciences, Monash University, Melbourne, Australia.
81. Department of Neuroscience, Central Clinical School, Monash University, Melbourne, Australia.

82. College of Health and Human Sciences, Colorado State University, Fort Collins, CO.
83. Molecular Cellular Integrative Neuroscience Program, Colorado State University, Fort Collins, CO.
84. Department of Chemical and Biological Physics, Weizmann Institute of Science, Rehovot, Israel.
85. University Health Network, Toronto, Ontario, Canada.
86. Centre for Research in Neurodegenerative Diseases, University of Toronto, Toronto, Ontario, Canada.
87. Department of Imaging Physics, The University of Texas M. D. Anderson Cancer Center, Houston, TX.
88. Department of Psychiatry, Harvard Medical School, Boston, MA.
89. Massachusetts General Hospital, Boston, MA.
90. Department of Psychology and School of Data Science, University of Virginia, Charlottesville, VA.
91. Department of Radiology, University of Pennsylvania, PA.
92. Cohen Veterans Bioscience, New York City, NY.
93. Division of Neuropsychiatry and Neuromodulation, Massachusetts General Hospital and Harvard Medical School, Boston, MA.
94. Department of Physical Medicine & Rehabilitation, Virginia Commonwealth University, Richmond, VA.
95. Richmond Veterans Affairs (VA) Medical Center, Central Virginia VA Health Care System, Richmond, VA.
96. Department of Psychology, Georgia State University, Atlanta, GA.
97. Department of Neurology, University of Utah, Salt Lake City, UT.
98. Department of Psychology, Alberta Children's Hospital Research Institute, and the Hotchkiss Brain Institute, University of Calgary, Calgary, Canada.
99. Spaulding Rehabilitation Hospital, Boston, MA.
100. Department of Physical Medicine & Rehabilitation, Harvard Medical School, Boston, MA.
101. Department of Radiology, Stanford, CA.
102. Departments of Pediatrics, and Radiology, University of Utah, Salt Lake City, UT.
103. Departments of Pediatrics, Neurology, and Neuroscience, University of Florida, Gainesville, FL.
104. Departments of Neurology, Pediatrics, Psychiatry, Radiology, Engineering, and Ophthalmology, USC, Los Angeles, CA.
105. Department of Psychology, Penn State University, State College, PA.
106. Department of Neurology, Hershey Medical Center, State College, PA.

**Abstract**

An extensive library of symptom inventories has been developed over time to measure clinical symptoms, but this variety has led to several long standing issues. Most notably, results drawn from different settings and studies are not comparable, which limits reproducibility. Here, we present an artificial intelligence (AI) approach using semantic textual similarity (STS) to link symptoms and scores across previously incongruous symptom inventories. We tested the ability of four pre-trained STS models to screen thousands of symptom description pairs for related content - a challenging task typically requiring expert panels. Models were tasked to predict symptom severity across four different inventories for 6,607 participants drawn from 16 international data sources. The STS approach achieved 74.8% accuracy across five tasks, outperforming other models tested. This work suggests that incorporating contextual, semantic information can assist expert decision-making processes, yielding gains for both general and disease-specific clinical assessment.

**Key points:**

- We illustrate an AI approach using semantic textual similarity to link clinical terminologies and scores across previously incongruous symptom inventories.
- The approach significantly outperformed other benchmark models when tasked to predict scores across dually administered inventories.
- The model is provided as a free online tool enabling the comparison of scores across distinct symptom inventories.
- This work demonstrates the value of integrating AI into ongoing efforts to harmonise clinical and research data across settings and studies, which is needed to address high priority research questions requiring large sample sizes.

# Introduction

Self-reported symptom inventories are essential tools across clinical and research settings.[1–6] For example, standard clinical practice for patients with mild traumatic brain injury (TBI) enacts a symptom-based approach to direct treatment, but documenting heterogenous symptoms is a complex process.[7,8] As a result, a wide variety of self-reported symptom inventories have been developed over time, each with distinct items, phrasings, use cases, and reference periods.[2–6] To give one example, a recent report found at least nine different symptom inventories are used by athletic trainers to assess sports-related concussion. [9] Beyond TBI, hundreds of distinct symptom assessments are used across clinical and research settings. This seriously undermines reproducibility and our capacity to synthesise findings drawn from distinct sources.[10]

To alleviate this problem, clinicians who regularly use these instruments form expert panels to identify similar symptomatology across different instruments and co-calibrate results.[11–14] This expert panel-based approach is labour-intensive, and its inherent subjectivity can introduce noise and bias.[15] Despite expert panel efforts to identify standard definitions and controlled terminologies, only modest reductions in the variety of instruments have been achieved.[16,17] Meanwhile, the number of comparisons needed is large,[18] while efforts to harmonise inventories are slow, costly, and can involve hundreds of experts.[19]

Supplementing human expertise with artificial intelligence (AI) tools has the potential to enhance diagnosis, reduce disease-related burden, and automate laborious processes.[20,21,22,23] Beyond diagnosis, AI has the potential to assist with complex clinical and research processes where intuition and domain expertise are required.[24] Here, we report an international collaboration between neuroscientists, clinicians, and AI experts to robustly link symptom inventories using semantic textual similarity (STS). Our approach leverages the meaningful relationship between descriptions of symptoms to identify related content.[25]

We demonstrate how pre-trained Natural Language Processing (NLP) models can offer a rapid and accurate means to quantify the relationships between thousands of symptoms across measures. We focus on four inventories, two that are commonly used to assess general symptoms: The Brief Symptom Inventory–18, and The Symptom Checklist-90-Revised, as well as two inventories specific to brain injury: The Neurobehavioral Symptom Inventory, and The Rivermead Post-Concussion Symptoms Questionnaire.[2–6] We prioritised TBI because it is frequently clinically assessed using self-reported symptom inventories.[7] In addition, TBI research lacks the consensus within the field that research into other conditions have achieved, perhaps due to the heterogeneity of TBI clinical presentation and the multitude of existing TBI instruments.[26]

We hypothesised that an STS approach incorporating contextual semantic information would outperform traditional and machine learning models when tasked to predict participant scores on symptoms across different inventories. This hypothesis was based on our observation that the application of meaningful clinical intuition (for which STS is a potential proxy) can sometimes better predict and explain trends in noisy medical data than pure learning models. Following AI safety and reporting requirements,[27,28] we tested this hypothesis by assessing the ability of STS-linked items to estimate cross-inventory symptom scores for patients who were dually assessed on different inventories (n=2,056). The resulting analysis pipeline is available as a free online tool [`github.com/ShashanKV98/symptom-inventories`] that can convert scores across previously incompatible symptom inventories . This study presents a tangible example of how AI tools may help to mitigate long standing health data compatibility issues, not only for specific conditions like TBI, but also for general health assessments.

## Methods

### *Inventory Data Sources*

This secondary mega-analysis[29] study petitioned collaborators for item-level data, drawing from the Enhancing NeuroImaging Genetics through Meta-Analysis (ENIGMA) Brain Injury working group,[30,31] and the Long-term Impact of Military-relevant Brain Injury Consortium—Chronic Effects of Neurotrauma Consortium (LIMBIC-CENC).[32] We also included public data from the Federal Interagency Traumatic Brain Injury Research Informatics System (FITBIR).[33] We obtained 16 datasets that included different combinations of symptom inventories (see **Supplementary Table S1**). Data quality and consistency were confirmed during discussions among authors who collected the primary data. The University of Utah provided overall institutional review board (IRB) approvals and data use agreements. All self-reported measures were completed or administered in English.

### *Measures*

Comprehensive details of the four self-reported symptom inventories are provided in **Supplementary Note 1**. Briefly, we assessed two TBI related inventories: the Neurobehavioral Symptom Inventory (NSI[4]) and the Rivermead Post-Concussion Symptoms Questionnaire (RPQ[2,3]), and two general symptom inventories commonly used for TBI: The 18-item Brief Symptom Inventory - 18 Item version (BSI-18[5]), and The Symptom Checklist-90-Revised (SCL-90-R[6]).

The 22-item NSI[4] evaluates cognitive, somatic, and emotional symptoms commonly experienced by adults following a brain injury, including headache, dizziness, irritability, and difficulty concentrating. Symptom frequency and severity are measured on a five-point Likert-Scale, where the respondent indicates the degree to which they were disturbed by each symptom over the past two weeks from 0 (*None*) to 4 (*Very severe*).

The 16-item RPQ[2,3] measures the presence and severity of commonly reported TBI symptoms, across somatic, cognitive, and emotional domains, including headache, dizziness, fatigue, irritability, and concentration difficulties. Using a five-point Likert-Scale from 0 (*not experienced*) to 4 (*severe problem*), respondents are instructed to rate the severity of each symptom experienced within the last 24 hours, relative to their experience of the symptom before injury.

The 18-item BSI-18[5] is a shortened version of the Symptom Checklist-90-Revised (SCL-90-R) and the original 53-item BSI. The BSI-18 is designed to efficiently and broadly assess psychological symptoms in both healthy and patient populations. The BSI-18 consists of items rated on a five-point scale, ranging from 0 (*not at all*) to 4 (*extremely*), indicating how much the problems distressed or bothered respondents over the past seven days.[5]

The Symptom Checklist-90-Revised (SCL-90-R[6]) is a widely used self-report measure designed to assess the presence, severity, and frequency of 90 broad psychological symptoms and measures of emotional distress, and like the BSI-18, is not just specific to TBI. Respondents rate each item on a five-point scale from 0 (*Not at all*) to 4 (*extremely*), indicating how much they have been bothered or distressed by the symptom over the past week.

**Semantic Textual Similarity**

This study used the similarity of question-level text descriptions to identify related items across inventories. This is a challenging task because symptom descriptions can have similar meanings, yet share no words in common. For example, "*Vision problems," "blurring," "trouble seeing,*" and "*Light sensitivity*" all relate to vision/ocular symptoms, but they do not include the same words. Advancements in NLP have yielded tools that can rapidly score the semantic similarity of text, such as transformer models trained on a large corpus of text to encode and represent text strings within an embedded feature space.[37-40] This approach has two main advantages: 1) sentences of arbitrary length are converted to an embedded feature vector of prespecified length. This means different lengths of text can be directly compared as representations of prespecified length, and 2) sentences closer in the embedding space are more semantically similar, so the distance between feature vectors measures the meaningful similarity of text.

**AI Safety and Reporting Criteria**

AI modelling and usage were conducted in accordance with guidelines and quality criteria for AI-based research.[27,28] To protect patient privacy, models that were additionally trained on sensitive clinical operations data were not published online, per recommendations for safe use of large language models.[34] All models were developed as research tools and should not be used for decision-making in individual clinical cases. We followed eight recommended guidelines for AI study reporting, which included:

1. *Data sources:* All data sources are outlined in **Supplementary Table S1**.

2. *Data pre-processing:* Raw, unadjusted scores were used. Symptom inventories with one or more scores missing were excluded.

3. *Partitioning:* The same 50/50 train-test splitting of participants was performed for all models.

4. *Disjointness:* Participant data was fully disjoint to avoid duplicates across training and test data. Any repeated measurements over time for the same participants of the same inventory were dropped. When learning to convert numeric scores from one inventory to another, only one inventory type was permitted in the training data for all participants at a time, and only one test-inventory item was permitted as the target.

5. *Models and Training*: Four STS models were evaluated, each pre-trained on different corpuses of general and medical text: **(a)** For the base model (*MiniLMBERT*), a pre-trained Bidirectional Encoder Representations from Transformers (BERT) model was used. *MiniLMBERT* distilled the self-attention module of the last transformer layer of a large transformer[35] trained on several million sentence pairs.[36] This model is publicly available online.[37] Three other models pre-trained on biomedical and clinical text were also evaluated: **(b)** *ClinicalCovidBERT:* A publicly available model pre-trained on the CORD-19 medical dataset.[38] **(c)** *VAClinicalDocsBERT:* A clinically trained model that used *ClinicalCovidBERT* as a base, but with additional pre-training on 500,000 generic clinical operational documents from the Veterans Affair (VA) Corporate Data Warehouse (CDW[39]), including admissions and discharge summaries. **(d)** *VAMetadataBERT:* A medically trained model that used *ClinicalCovidBERT* as a base, with additional training on 1.5 million text strings of clinical lab names, medication names, and document titles from the VA CDW.

*6. Hyperparameters and Tuning:* To improve reproducibility, no fine tuning was performed. We did not change any of the weights of models at any stage to tune to symptom inventory content.

*7. Model Selection:* The four BERT models were evaluated by comparing their relative performance at the task of correctly converting scores across inventories. Given scores on one set of inventory items, the task was to estimate all scores on another, fully distinct, set of inventory items. The ground truth for this problem was two dually administered inventories per person. Performance was only evaluated on held-out test participants.

*8. Model Metrics.* The primary metric in study was cosine similarity, S, which measures the semantic similarity of two symptom descriptions in the range 0 - 1. A value of 0 indicates the symptom descriptions share no meaningful similarity, while 1 means they have identical meanings. Model prediction performance was measured using mean absolute error (MAE), binary accuracy, and multinomial accuracy. Multinomial, or exact match accuracy (EMA), was defined as the percentage of estimated scores that equaled the correct symptom severity on a five-point Likert scale. Although accuracies near 50% normally indicate poor performance, the random guess accuracy was 20% for EMA on a five-point scale. EMA was our preferred metric since it measures accuracy on the true scale, and also strongly correlates with MAE (**see Supplementary Figure S1**).

**Crosswalk Model**

A 'crosswalk' refers to the process of relating items of one inventory to another. A conceptual overview of the STS model process is shown in **Figure 1**. First, the pre-trained BERT transformer scored the similarity of symptom descriptions across inventories (**Figure 1a**). For each item, its most similar item descriptions were found on other inventories.

The second step adjusted for differences in scale response across items (**Figure 1b**). For example, on the BSI-18, 'Mild' is defined as the second option on the five-point Likert scale. In contrast, 'Mild' is the third point on the Likert scale for the RPQ. Therefore, different raw scores imply the same symptom severity level. A percentile sampling approach was used to mitigate these differences (see **Supplemental Figure S2**). If items had no single close analogue on other inventories, multiple items were used for prediction (**Figure 1c**). After model construction, performance was assessed by comparing estimated and actual inventory scores for dually administered assessments (**Figure 1d**).

**Statistical Analysis**

Analysis was performed in Python 3. Validation data required no covariate adjustment, as the same set of individuals were dually administered the same two inventories. Chi-squared tests were used to assess categorical variables and *t*-tests were used to compare continuous variables. The *sentence-transformers*[37] Python package was used for text embedding, while the *statsmodels*[40] and *scikit-learn* packages[41] were used to construct linear and machine learning models, respectively.

For prediction, we define *A* and *B* to be two dually administered assessments. The goal was to predict a single item in *B*, named *b*, given all items in *A*. Various strategies were explored to predict scores across inventories. Overall, a nearest-neighbour (NN) approach was implemented that selected only the most semantically similar item in *A*, $a = \text{argmax}[S(A,b)]$, to predict *b*, excluding all other scores.

# Results

## Data Summary

**Table 1** summarises characteristics of the cohort (n=6,607). The cohort showed good representation across age, sex, education level, race, ethnicity, and TBI status. In terms of injury severity, 1,159 participants (17.5%) were controls with no history of TBI, 5,400 participants (81.7%) had a history of mild TBI, and just 48 participants (0.7%) had a history of moderate/severe TBI. Across all participants, the median age was 29 years old, with an interquartile range of 20-43 years, and 29.4% were female. Using appropriate scoring schema, the means (and standard deviations) of total scores were BSI-18: 8.86 (10.52), RPQ: 17.49 (14.83), SCL-90-R: 70.5 (67.21), and NSI: 25.5 (16.95). An overlapping sample of the same participants (n=2,056) was administered both the BSI and RPQ; these included 286 controls.

## Semantic Text Similarity

Initial analysis was performed using the general language model, *MiniMLBERT*. Illustrative examples of *MiniMLBERT* symptom similarities are shown in **Table 2** for symptom pair comparisons. **Figure 2** shows a stem plot of cosine similarities for one RPQ symptom, 'Nausea and/or vomiting,' compared to all BSI-18 symptom descriptions. Most of the 18 BSI-18 items were classified as conceptually unrelated to the item, but 'Nausea or upset stomach' was strongly related. As these two items had maximum similarity, they formed one cross-inventory pair. Conversely, the model did not link unrelated items, even if they contained overlapping words (see **Supplementary Note 2**).

To extend beyond single examples, **Figure 3** shows the similarity scores of all items across all inventories (NSI, SCL-90-R, BSI-18, and RPQ), sorted and colour-coded by cross-inventory comparison. The SCL→BSI comparison had 18 near-identical item pairs (*cyan stars, top*). This means that the BSI-18 was effectively a semantic subset of the SCL-90-R, as would be expected given the BSI-18 uses items from the SCL-90 R. In this way, STS can rapidly screen for closely related items across inventories, regardless of whether these relationships are established or not in the literature.

To assess the semantic similarity of different inventories in aggregate, the distribution of closest-pair cosine similarities was found for each inventory pair (**Figure 4**). Overall, 41.7% of the closest pairs were $S>0.6$, indicating that many single symptoms had close analogues in other inventories. When considering aggregate similarity across inventories, directionality matters (i.e., A→B vs. B→A). The NSI and RPQ, both TBI-related inventories, contained similar content, whereas the NSI and BSI-18 showed relatively low median similarity.

**Inventory Score Prediction**

Before implementing cross-inventory models, within-inventory score prediction was implemented to assess the ability to convert scores in the absence of cross-inventory effects (**Supplementary Figure S3**). Within-inventory models used data for all items in a given inventory (except one) as explanatory variables to estimate scores on the single, reserved item. As these experiments were not subject to any cross-inventory effects (e.g., differences in administration or scoring), they estimate an upper bound on accuracy free of cross-inventory effects. The average prediction accuracy was 57.7% for within-inventory estimation.

**Figure 5** shows the *cross*-inventory prediction accuracies for four different models: 1. Semantic Textual Similarity, 2. Linear regression, 3. Random Forest, and 4. Gradient Boosting. The data included n=2,056 individuals who were administered both the RPQ and BSI-18, split randomly into 50/50 test-train groups (designating n=1,028 held out test participants). Only RPQ items were used to estimate BSI item scores (RPQ→BSI-18, grey circles), and only BSI-18 items used to estimate RPQ item scores (BSI-18→RPQ, white circles). The STS model (blue line) achieved 54.3% accuracy when predicting scores across inventories, consistently outperforming the benchmark models across a wide range of symptoms, and reaching close to the estimated upper bound (57.7%).

**Model Performance**

Benchmark and AI models were evaluated by comparing their relative performance at cross-inventory symptom score conversion for five different multinomial and binary classification tasks (**Supplemental Table S2**). Overall, the *MiniLMBERT* model trained on a general corpus of text showed the highest accuracy, 74.8%, when averaged across all scenarios, consistently outperforming the three clinically pretrained STS transformers (72.9% - 73.8%). *MiniLMBERT* also achieved 54.3% on the more challenging multinomial EMA prediction task, outperforming all benchmark models (40.6% - 48.0%), and the three other medically pretrained transformers (52.0% - 53.2%). The symptom inventory conversion tool is available as a web interface (see **Supplementary Figure S4**).

To explore whether model efficacy varied across sex, *MiniLMBERT* performance was stratified for dually administered male (N=1,349) and female (N=707) participants. The model predicted female symptoms with 6% lower accuracy than male symptoms, equivalent to an effect size of $d = -0.43$ ($p<0.001$). Relatedly, only 14.7% of the training cohort were female. An age-stratified performance evaluation was conducted for two groups of dually administered participants; aged 65 years or above (N=190), and aged below 65 years (N=1,866). Interestingly, symptoms were more accurately predicted for the elderly group (EMA: 61.7%), than for those below 65 years of age (EMA: 53.4%, $p<0.001$).

## Discussion

Providers often use standard inventories for initial evaluation and tracking of symptoms for many health conditions. However, comparing results across distinct symptom inventories is challenging due to subtle differences in how symptoms are described, assessed, and conceptualised. These differences confound the aggregation of data and findings across clinical and research settings, and also limit the comparison between historical and current studies. To address this problem, this study reported a novel application of AI language models to rapidly and accurately link items and scores across self-reported symptom inventories.

We tested four semantic symptom-linking models on data for thousands of individuals who each completed two different symptom inventories. Overall, a deep learning model trained on a general text corpus showed the highest accuracy, which confirmed our hypothesis. The superior performance of the generic language model over clinically pretrained models is consistent with the straightforward language inventories used to describe symptoms.

One issue when comparing symptom inventories is that their content can overlap. For example the BSI-18 is a short version contained in the SCL-90. Initially considered a potential challenge for the work, the existence of direct analogues across inventories was valuable because they provided identical semantic ground truths across inventories. This facilitated the direct observation of the effects of different inventory scoring schemes and scales for otherwise identical items.

This study also offers useful insights into the nature of TBI-related symptomatology and measurement. Many studies conduct multiple inventory assessments to more completely capture a wide range of potential patient experiences. Semantic insights could help to guide and optimise the selection of complementary instruments. Across inventories, about two-thirds of the NSI and RPQ symptoms were strongly related, confirming that they were semantically similar assessments, which was anticipated since both assess TBI. By contrast, those wishing to pair a general and TBI specific inventory could consider the NSI and BSI-18, as they had lower average similarity than other inventory pairs. Beyond existing inventories, deep learning text similarity paradigms might also be able to assist in the development of new, abbreviated questionnaires that more precisely assess distress, and one could imagine a synthetic superscale that draws semantically from all inventories.

Although 'harmonisation' commonly refers to data aggregation and cleaning, true data harmonisation aims to minimise unwanted measurement variations while preserving the underlying meaning of the measures of interest. In the course of developing an AI pipeline for cross-walking across symptom inventories, we observed that the STS model did not always link items with the highest empirical correlation on scores. Instead, it detected and leveraged subtle relationships between symptom phrasings, and in doing so, exceeded the performance of other empirically trained linear and machine learning models. The current finding that the similarity of text describing symptoms was generally more useful than training on empirical data is surprising. Perhaps the descriptive similarity of study measures themselves could be used in other health domains to outperform pure learning models.

Many studies using AI in medicine have demonstrated impressive gains in diagnostic accuracy, but the diagnostic labels needed to train AI models are often assigned using clinical evaluation tools with long-standing data compatibility issues. This study leveraged AI to address this more fundamental decision-focused task - the harmonisation of clinical measurements. If AI can be used to improve the quality of tools that assign training data labels, then it may be possible to achieve further, untapped gains in accuracy across a range of health-related deep learning tasks.

**Strengths and Limitations**

Strengths of this study include a large aggregated sample drawn from 16 data sources, high quality dually administered test data, evaluation of multiple AI models trained on different medical and general text sources, and detailed investigations of both disease-specific and general symptom inventories. Other strengths include a close collaboration between AI and clinical experts that ensured patient safety, privacy, and careful adherence to recommended AI reporting criteria, and both data and code are made available.

There are some limitations of the current study worth noting. First, three or more inventories per participant were unavailable. We also did not adjust for symptom validity. Three of the inventories use distinct reference time frames, and differences in administration were not considered. However, since the method was nearly as accurate when estimating scores across, compared to within, inventories, cross-assessment effects were largely mitigated. Fourth, the data were drawn from 18 English language sources, including both military and civilian datasets. Therefore, the findings may not generalise to specific populations. However, insofar as this was tested, stratifying the results by age and sex showed only modest variations in performance.

Inventories do not capture all elements of personal experience, and some may even systematically screen out or inadequately capture meaning. The extent to which crosswalk tools incur related loss of information regarding the patient's experience should be studied. Only inventories with five point scales were used, but there is no reason why this approach could not be extended to inventories with different numeric scales. Nevertheless, this study utilised one of the largest samples of symptom inventory data in TBI yet assembled and focused upon the most widely used and recommended measures in the field.[42–44]

## Author contributions

Original data were collected by Drs. Troyanskaya, Newsome, Morey, Tate, Walker, and Wilde. Drs. Kennedy, Dennis, Wilde, Tate, Hillary, Dams-O'Connor, Lindsey, and Liebel conceived of the project. Drs. Dennis and Lindsey compiled and cleaned the data. Dr. Kennedy, Mr. Vadlamani, Dr. Peterson, and Mr. Agarwal completed the processing and analysis. Dr. Kennedy wrote the manuscript, and all authors reviewed, edited, and approved the manuscript.

## Data availability

We included 16 different sources of data in this study. Of these, six data sources are freely available online as part of the Federal Interagency Traumatic Brain Injury Research Informatics System (FITBIR) data repository, hosted at `https://fitbir.nih.gov`. The other ten datasets are available from the corresponding author on reasonable request, pending IRB approval for dissemination of data, and additional institutional approval of data use and access.

## Code availability

Symptom inventory conversion codes are available at: `github.com/ShashanKV98/symptom-inventories`.

## Competing Interests

None of the authors have competing interests relevant to this manuscript. Neda Jahanshad and Paul Thompson received a research grant from Biogen, Inc., for research unrelated to this manuscript. Timothy Meier receives compensation as a member of the Clinical and Scientific Advisory Board for Quadrant Biosciences, Inc. Virginia Newcome holds a grant from Roche Pharmaceuticals for a project unrelated to this manuscript. Alexander Olsen is a co-founder and owner of Nordic Brain Tech AS.

## Acknowledgements

This work was supported by R61NS120249 to ELD, EAW, FGH, and DFT. The views expressed in this article are those of the author(s) and do not reflect the official policy of the Department of Army/Navy/Air Force, Department of Defense, or U.S. Government.

# References

1. Wallace, W. *et al.* The diagnostic and triage accuracy of digital and online symptom checker tools: a systematic review. *NPJ Digit Med* **5**, 118 (2022).
2. King, N. S., Crawford, S., Wenden, F. J., Moss, N. E. & Wade, D. T. The Rivermead Post Concussion Symptoms Questionnaire: a measure of symptoms commonly experienced after head injury and its reliability. *J. Neurol.* **242**, 587–592 (1995).
3. Potter, S., Leigh, E., Wade, D. & Fleminger, S. The Rivermead Post Concussion Symptoms Questionnaire: a confirmatory factor analysis. *J. Neurol.* **253**, 1603–1614 (2006).
4. Cicerone, K. D. & Kalmar, K. Persistent postconcussion syndrome: The structure of subjective complaints after mild traumatic brain injury. *J. Head Trauma Rehabil.* **10**, 1–17.
5. Derogatis, L. R. Brief symptom inventory 18. https://www.researchgate.net/profile/Rachel-Bachner-Melman/post/Can_the_SCL-10R_be_used_to_assess_risk_for_multiple_mental_health_conditions/attachment/5fbfa50b2f079e0001be051a/AS%3A962105134022658%401606395147689/download/bsi+info.docx.
6. Derogatis, L. R. & Unger, R. Symptom Checklist-90-Revised. *The Corsini Encyclopedia of Psychology* (2010) doi:10.1002/9780470479216.corpsy0970.
7. Maas, A. I. R. *et al.* Traumatic brain injury: progress and challenges in prevention, clinical care, and research. *Lancet Neurol.* **21**, 1004–1060 (2022).
8. Karaliute, M. *et al.* Methodology Matters: Comparing Approaches for Defining Persistent Symptoms after Mild Traumatic Brain Injury. *Neurotrauma Rep* **2**, 603–617 (2021).
9. Lempke, L. B., Schmidt, J. D. & Lynall, R. C. Athletic Trainers' Concussion-Assessment and Concussion-Management Practices: An Update. *J. Athl. Train.* **55**, 17–26 (2020).
10. Rosenbusch, H., Wanders, F. & Pit, I. L. The Semantic Scale Network: An online tool to detect semantic overlap of psychological scales and prevent scale redundancies. *Psychol. Methods* **25**, 380–392 (2020).
11. Scollard, P. *et al.* Ceiling effects and differential measurement precision across calibrated cognitive scores in the Framingham Study. *Neuropsychology* **37**, 383–397 (2023).
12. Nance, R. M. *et al.* Co-calibration of two self-reported measures of adherence to antiretroviral therapy. *AIDS Care* **29**, 464–468 (2017).
13. Mukherjee, S. *et al.* Cognitive domain harmonization and cocalibration in studies of older adults. *Neuropsychology* **37**, 409–423 (2023).
14. Boccardi, M. *et al.* Harmonizing neuropsychological assessment for mild neurocognitive disorders in Europe. *Alzheimers. Dement.* **18**, 29–42 (2022).
15. Pinkham, A. E. *et al.* The social cognition psychometric evaluation study: results of the expert survey and RAND panel. *Schizophr. Bull.* **40**, 813–823 (2014).
16. Dams-O'Connor, K. *et al.* Alzheimer's Disease-Related Dementias Summit 2022: National Research Priorities for the Investigation of Post-Traumatic Brain Injury Alzheimer's Disease and Related Dementias. *J. Neurotrauma* **40**, 1512–1523 (2023).
17. Meeuws, S. *et al.* Common Data Elements: Critical Assessment of Harmonization between Current Multi-Center Traumatic Brain Injury Studies. *J. Neurotrauma* **37**, 1283–1290 (2020).
18. Polinder, S., Haagsma, J. A., van Klaveren, D., Steyerberg, E. W. & van Beeck, E. F. Health-related quality of life after TBI: a systematic review of study design, instruments, measurement properties, and outcome. *Popul. Health Metr.* **13**, 4 (2015).
19. Brady, K. J. S. *et al.* Establishing Crosswalks Between Common Measures of Burnout in US Physicians. *J. Gen. Intern. Med.* **37**, 777–784 (2022).
20. He, K., Zhang, X., Ren, S. & Sun, J. Deep residual learning for image recognition. in *2016 IEEE Conference on Computer Vision and Pattern Recognition (CVPR)* 770–778 (IEEE, 2016).
21. Lu, B. *et al.* A practical Alzheimer's disease classifier via brain imaging-based deep learning on 85,721 samples. *Journal of Big Data* **9**, 1–22 (2022).
22. Liu, Y. *et al.* RoBERTa: A Robustly Optimized BERT Pretraining Approach. *arXiv [cs.CL]* (2019).
23. Wang, A. *et al.* GLUE: A Multi-Task Benchmark and Analysis Platform for Natural Language

Understanding. *arXiv [cs.CL]* (2018).
24. Rajpurkar, P., Chen, E., Banerjee, O. & Topol, E. J. AI in health and medicine. *Nat. Med.* **28**, 31–38 (2022).
25. Chandrasekaran, D. & Mago, V. Evolution of Semantic Similarity—A Survey. *ACM Comput. Surv.* **54**, 1–37 (2021).
26. Prince, C. & Bruhns, M. E. Evaluation and Treatment of Mild Traumatic Brain Injury: The Role of Neuropsychology. *Brain Sci* **7**, (2017).
27. Mongan, J., Moy, L. & Kahn, C. E., Jr. Checklist for Artificial Intelligence in Medical Imaging (CLAIM): A Guide for Authors and Reviewers. *Radiol Artif Intell* **2**, e200029 (2020).
28. Roberts, M. *et al.* Common pitfalls and recommendations for using machine learning to detect and prognosticate for COVID-19 using chest radiographs and CT scans. *Nature Machine Intelligence* **3**, 199–217 (2021).
29. Eisenhauer, J. G. Meta‐analysis and mega‐analysis: A simple introduction. *Teach. Stat.* **43**, 21–27 (2021).
30. Dennis, E. L. *et al.* ENIGMA brain injury: Framework, challenges, and opportunities. *Human Brain Mapping* Preprint at https://doi.org/10.1002/hbm.25046 (2020).
31. Wilde, E. A., Dennis, E. L. & Tate, D. F. The ENIGMA Brain Injury Working Group: Approach, Challenges, and Potential Benefits. *Brain Imaging Behav.* **Under Review**, (2019).
32. Cifu, D. X. & Dixon, K. J. Chronic effects of neurotrauma consortium. *Brain Inj.* **30**, 1397–1398 (2016).
33. Home. https://fitbir.nih.gov/.
34. Meskó, B. & Topol, E. J. The imperative for regulatory oversight of large language models (or generative AI) in healthcare. *NPJ Digit Med* **6**, 120 (2023).
35. Vaswani, A. *et al.* Attention is all you need. *Adv. Neural Inf. Process. Syst.* **30**, (2017).
36. Wang, W. *et al.* MiniLM: Deep Self-Attention Distillation for Task-Agnostic Compression of Pre-Trained Transformers. *arXiv [cs.CL]* (2020).
37. Reimers, N. & Gurevych, I. Sentence-BERT: Sentence Embeddings using Siamese BERT-Networks. *arXiv [cs.CL]* (2019).
38. Chapman, A. B. *et al.* A Natural Language Processing System for National COVID-19 Surveillance in the US Department of Veterans Affairs. (2020).
39. Health services research & Development. https://www.hsrd.research.va.gov/.
40. Seabold, S. & Perktold, J. Econometric and statistical modeling with Python skipper seabold 1 1. *Proc 9th Python Sci Conf*.
41. Pedregosa, F. *et al.* Scikit-learn: Machine Learning in Python. *arXiv [cs.LG]* 2825–2830 (2012).
42. Wilde, E. A. *et al.* Recommendations for the use of common outcome measures in traumatic brain injury research. *Arch. Phys. Med. Rehabil.* **91**, 1650–1660.e17 (2010).
43. Hicks, R. *et al.* Progress in developing common data elements for traumatic brain injury research: version two--the end of the beginning. *J. Neurotrauma* **30**, 1852–1861 (2013).
44. Broglio, S. P. *et al.* National Institute of Neurological Disorders and Stroke and Department of Defense Sport-Related Concussion Common Data Elements Version 1.0 Recommendations. *J. Neurotrauma* **35**, 2776–2783 (2018).
45. Lynall, R. C. *et al.* Optimizing Concussion Care Seeking: The Influence of Previous Concussion Diagnosis Status on Baseline Assessment Outcomes. *Am. J. Sports Med.* **50**, 3406–3416 (2022).
46. O'Connor, K. L. *et al.* Descriptive Analysis of a Baseline Concussion Battery Among U.S. Service Academy Members: Results from the Concussion Assessment, Research, and Education (CARE) Consortium. *Mil. Med.* **183**, e580–e590 (2018).
47. Katz, B. P. *et al.* Baseline Performance of NCAA Athletes on a Concussion Assessment Battery: A Report from the CARE Consortium. *Sports Med.* **48**, 1971–1985 (2018).
48. Morey, R. A. *et al.* Effects of chronic mild traumatic brain injury on white matter integrity in Iraq and Afghanistan war veterans. *Hum. Brain Mapp.* **34**, 2986–2999 (2013).
49. Clausen, A. N. *et al.* Combat exposure, posttraumatic stress disorder, and head injuries differentially relate to alterations in cortical thickness in military Veterans.

*Neuropsychopharmacology* vol. 45 491–498 Preprint at https://doi.org/10.1038/s41386-019-0539-9 (2020).
50. McMahon, P. *et al.* Symptomatology and functional outcome in mild traumatic brain injury: results from the prospective TRACK-TBI study. *J. Neurotrauma* **31**, 26–33 (2014).
51. Levin, H. S. *et al.* Association of Sex and Age With Mild Traumatic Brain Injury-Related Symptoms: A TRACK-TBI Study. *JAMA Netw Open* **4**, e213046 (2021).
52. Nelson, L. D. *et al.* Functional Recovery, Symptoms, and Quality of Life 1 to 5 Years After Traumatic Brain Injury. *JAMA Netw Open* **6**, e233660 (2023).
53. Cnossen, M. C. *et al.* Development of a Prediction Model for Post-Concussive Symptoms following Mild Traumatic Brain Injury: A TRACK-TBI Pilot Study. *J. Neurotrauma* **34**, 2396–2409 (2017).
54. Krieger, D. *et al.* MEG-Derived Symptom-Sensitive Biomarkers with Long-Term Test-Retest Reliability. *Diagnostics (Basel)* **12**, (2021).
55. Zhou, Y. *et al.* Mild traumatic brain injury: longitudinal regional brain volume changes. *Radiology* **267**, 880–890 (2013).
56. O'Neil, M. E. *et al.* Associations Among PTSD and Postconcussive Symptoms in the Long-Term Impact of Military-Relevant Brain Injury Consortium-Chronic Effects of Neurotrauma Consortium Prospective, Longitudinal Study Cohort. *J. Head Trauma Rehabil.* **36**, E363–E372 (2021).
57. Cifu, D. X. Clinical research findings from the long-term impact of military-relevant brain injury consortium-Chronic Effects of Neurotrauma Consortium (LIMBIC-CENC) 2013-2021. *Brain Inj.* **36**, 587–597 (2022).
58. Vanderploeg, R. D. *et al.* Predicting treatment response to cognitive rehabilitation in military service members with mild traumatic brain injury. *Rehabil. Psychol.* **63**, 194–204 (2018).
59. Troyanskaya, M. *et al.* Risk factors for decline in cognitive performance following deployment-related mild traumatic brain injury: A preliminary report. *Neurocase* **27**, 457–461 (2021).
60. Fischer, B. L. *et al.* Neural activation during response inhibition differentiates blast from mechanical causes of mild to moderate traumatic brain injury. *J. Neurotrauma* **31**, 169–179 (2014).
61. Newsome, M. R. *et al.* Disruption of caudate working memory activation in chronic blast-related traumatic brain injury. *Neuroimage Clin* **8**, 543–553 (2015).
62. Levin, H. S. *et al.* Diffusion tensor imaging of mild to moderate blast-related traumatic brain injury and its sequelae. *J. Neurotrauma* **27**, 683–694 (2010).
63. Weiner, M. W. *et al.* Effects of traumatic brain injury and posttraumatic stress disorder on development of Alzheimer's disease in Vietnam Veterans using the Alzheimer's Disease Neuroimaging Initiative: Preliminary Report. *Alzheimers. Dement.* **3**, 177–188 (2017).
64. Weiner, M. W. *et al.* Effects of traumatic brain injury and posttraumatic stress disorder on Alzheimer's disease in veterans, using the Alzheimer's Disease Neuroimaging Initiative. *Alzheimers. Dement.* **10**, S226–35 (2014).
65. King, P. R. *et al.* Psychometric study of the Neurobehavioral Symptom Inventory. *J. Rehabil. Res. Dev.* **49**, 879–888 (2012).
66. Caplan, L. J. *et al.* The Structure of Postconcussive Symptoms in 3 US Military Samples. *J. Head Trauma Rehabil.* **25**, 447 (2010).
67. Vos, L. *et al.* Comparison of the Neurobehavioral Symptom Inventory and the Rivermead Postconcussion Symptoms Questionnaire. *Brain Inj.* **33**, 1165–1172 (2019).
68. Meterko, M. *et al.* Psychometric assessment of the Neurobehavioral Symptom Inventory-22: the structure of persistent postconcussive symptoms following deployment-related mild traumatic brain injury among veterans. *J. Head Trauma Rehabil.* **27**, 55–62 (2012).
69. Vanderploeg, R. D. *et al.* The structure of postconcussion symptoms on the Neurobehavioral Symptom Inventory: a comparison of alternative models. *J. Head Trauma Rehabil.* **30**, 1–11 (2015).

70. Benge, J. F., Pastorek, N. J. & Thornton, G. M. Postconcussive symptoms in OEF-OIF veterans: factor structure and impact of posttraumatic stress. *Rehabil. Psychol.* **54**, 270–278 (2009).
71. Menatti, A. R. R., Melinder, M. R. D. & Warren, S. L. Limited Prediction of Performance Validity Using Embedded Validity Scales of the Neurobehavioral Symptom Inventory in an mTBI Veteran Sample. *J. Head Trauma Rehabil.* **35**, E36–E42 (2020).
72. Silva, M. A. Review of the Neurobehavioral Symptom Inventory. *Rehabil. Psychol.* **66**, 170–182 (2021).
73. Soble, J. R. *et al.* Normative Data for the Neurobehavioral Symptom Inventory (NSI) and post-concussion symptom profiles among TBI, PTSD, and nonclinical samples. *Clin. Neuropsychol.* **28**, 614–632 (2014).
74. Belanger, H. G. *et al.* [Formula: see text]Interpreting change on the neurobehavioral symptom inventory and the PTSD checklist in military personnel. *Clin. Neuropsychol.* **30**, 1063–1073 (2016).
75. de Guise, E. *et al.* Usefulness of the rivermead postconcussion symptoms questionnaire and the trail-making test for outcome prediction in patients with mild traumatic brain injury. *Appl. Neuropsychol. Adult* **23**, 213–222 (2016).
76. Ingebrigtsen, T., Waterloo, K., Marup-Jensen, S., Attner, E. & Romner, B. Quantification of post-concussion symptoms 3 months after minor head injury in 100 consecutive patients. *J. Neurol.* **245**, 609–612 (1998).
77. Eyres, S., Carey, A., Gilworth, G., Neumann, V. & Tennant, A. Construct validity and reliability of the Rivermead Post-Concussion Symptoms Questionnaire. *Clin. Rehabil.* **19**, 878–887 (2005).
78. Lannsjö, M., Borg, J., Björklund, G., Af Geijerstam, J.-L. & Lundgren-Nilsson, A. Internal construct validity of the Rivermead Post-Concussion Symptoms Questionnaire. *J. Rehabil. Med.* **43**, 997–1002 (2011).
79. Akhavan Abiri, F. & Shairi, M. R. Validity and reliability of symptom checklist-90-revised (SCL-90-R) and brief symptom inventory-53 (BSI-53). *Clinical Psychology and Personality* **17**, 169–195 (2020).
80. Meachen, S.-J., Hanks, R. A., Millis, S. R. & Rapport, L. J. The reliability and validity of the brief symptom inventory-18 in persons with traumatic brain injury. *Arch. Phys. Med. Rehabil.* **89**, 958–965 (2008).
81. Prinz, U. *et al.* Comparative psychometric analyses of the SCL-90-R and its short versions in patients with affective disorders. *BMC Psychiatry* **13**, 104 (2013).
82. Vallejo, M. A., Mañanes, G., Isabel Comeche, M. A. & Díaz, M. I. Comparison between administration via Internet and paper-and-pencil administration of two clinical instruments: SCL-90-R and GHQ-28. *J. Behav. Ther. Exp. Psychiatry* **39**, 201–208 (2008).
83. Hildenbrand, A. K., Nicholls, E. G., Aggarwal, R., Brody-Bizar, E. & Daly, B. P. Symptom Checklist-90-Revised (SCL-90-R). *The Encyclopedia of Clinical Psychology* 1–5 (2015) doi:10.1002/9781118625392.wbecp495.

**Table 1. Descriptive characteristics for the cohort by measure.** BSI-18: Brief Symptom Inventory–18, RPQ: Rivermead Post Concussion Symptoms Questionnaire, SCL-90-R: Symptom Checklist-90-Revised, NSI: Neurobehavioral Symptom Inventory. † indicates overlapping n=2,056 administration.

|  |  | BSI-18† | RPQ† | SCL-90-R | NSI |
|---|---|---|---|---|---|
| **Sample size (n):** |  | 4,286 | 2,153 | 244 | 1,984 |
| **Age (years):** | Mean (std) | 29.3 (15.3) | 39.4 (16.3) | 59.3 (15.9) | 39.0 (9.7) |
|  | > 65 years | 4.0% | 8.1% | 54.9% | 0.7% |
| **Sex:** | Male | 63.1% | 66.2% | 89.3% | 83.7% |
|  | Female | 36.9% | 33.8% | 10.7% | 16.3% |
| **Education:** | High School or less | 2.9% | 5.6% | 3.7% | 0.3% |
|  | Some College | 80.5% | 61% | 50% | 59.4% |
|  | Graduate degree | 16.6% | 33.4% | 46.3% | 40.3% |
| **Race:** | Black | 19.1% | 16.3% | 7% | 18.4% |
|  | White | 77.1% | 78.7% | 91.8% | 72.5% |
|  | Other | 3.8% | 5.0% | 1.2% | 9.1% |
| **Ethnicity:** | Hispanic/Latino | 14.3% | 19.7% | 2.5% | 16.5% |
| **History of TBI of any severity** |  | 88.3% | 86.7% | 57.0% | 75.2% |
| **Measure Total Score (Std)** |  | 8.9 (10.5) | 17.5 (14.8) | 70.5 (67.2) | 25.5 (17.0) |

**Table 2: Example comparisons of symptoms ranked by semantic similarity.** Small changes have been made to protect instrumental integrity. Overlapping words are shown in bold.

| Symptom text # 1 | Symptom text # 2 | Cosine Similarity |
|---|---|---|
| **Numbness or tingling** on **parts of** my **body** | **Numbness or tingling** in **parts of body** | 0.94 |
| Feeling **dizzy** | Faintness or **dizziness** | 0.76 |
| Nausea | Upset stomach | 0.72 |
| Feeling shy or uneasy with the opposite sex | Nervousness or shakiness inside | 0.55 |
| Feeling nervous when you are left alone | Suddenly scared for no reason | 0.45 |
| Sensitivity to light or sound | Feeling blue | 0.32 |
| Poor coordination | Feeling weak in parts of your body | 0.31 |
| **Trouble** remembering things | **Trouble** getting your breath | 0.20 |
| Blaming yourself for things | Change in taste and/or smell | 0.03 |

**Figure 1: Schematic representation of the STS crosswalk pipeline. (a)** Transformers scored the similarity of all symptom description pairs across inventories. The most similar item in the inventory to be linked was found for each symptom description. **(b)** Empirical differences in the response distribution of pairs were corrected using a stochastic sampling approach. **(c)** Scores with insufficiently similar linking items were predicted using within-inventory estimation after converting similar item scores. **(d)** The estimated and actual inventory scores were compared per participant to measure accuracy.

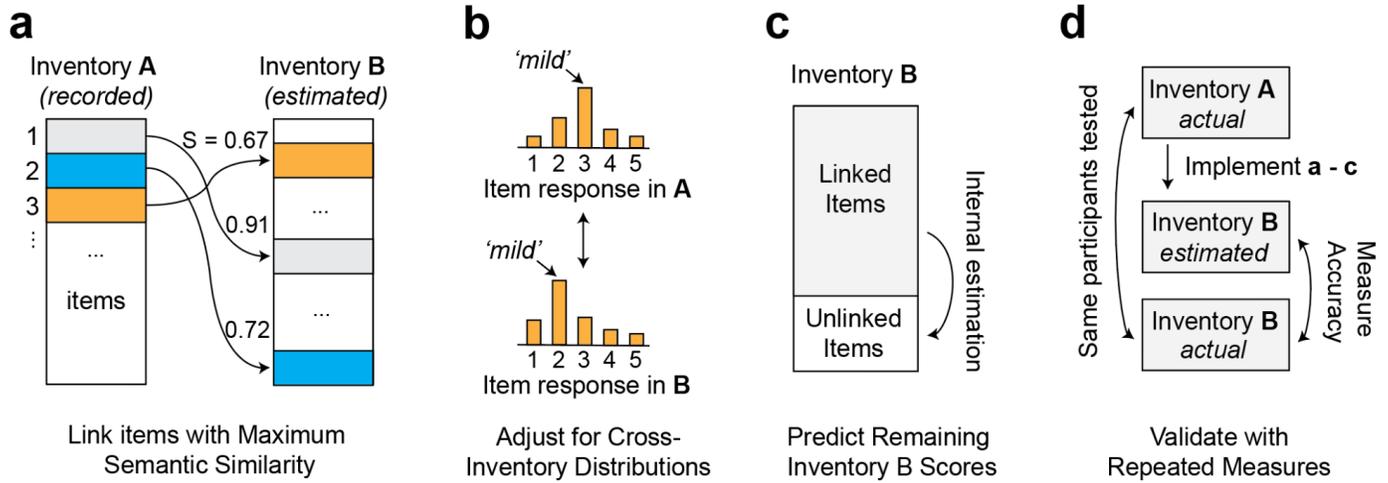

**Figure 2: Semantic text similarity (STS) of symptoms.** A stem plot shows the cosine similarity of the RPQ symptom 'Nausea and/or vomiting' with the 18 symptoms assessed by the BSI-18. BSI-18: Brief Symptom Inventory–18, RPQ: Rivermead Post Concussion Symptoms Questionnaire.

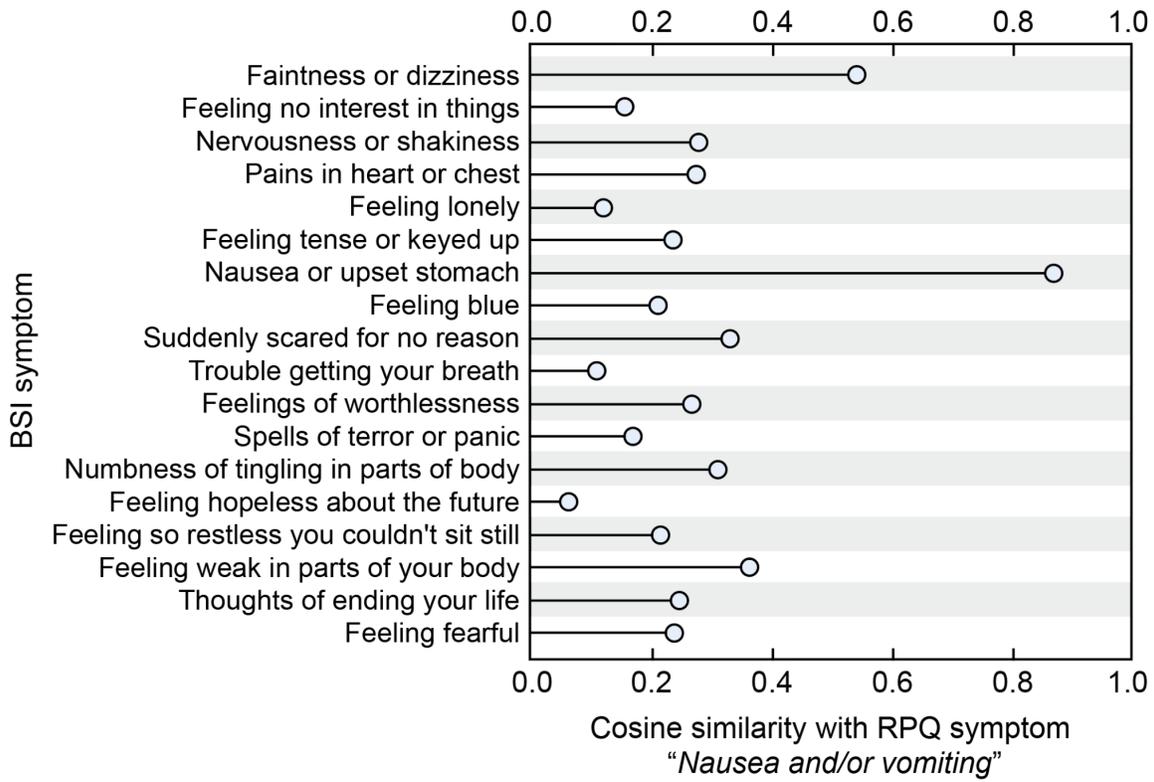

**Figure 3: Similarity of symptom pairs across all inventories.** The semantic text similarities for all items across the NSI, SCL-90-R, BSI-18, and RPQ are shown sorted, and colour-coded by inventory pair. Semantically related items are embedded in a large background of unrelated symptomatology; for example, the BSI-18 is a subset of the SCL-90-R, so there are 18 near-identical SCL-BSI similarity scores (*cyan stars, top*). BSI-18: Brief Symptom Inventory–18, RPQ: Rivermead Post Concussion Symptoms Questionnaire, SCL-90-R: Symptom Checklist-90-Revised, NSI: Neurobehavioral Symptom Inventory.

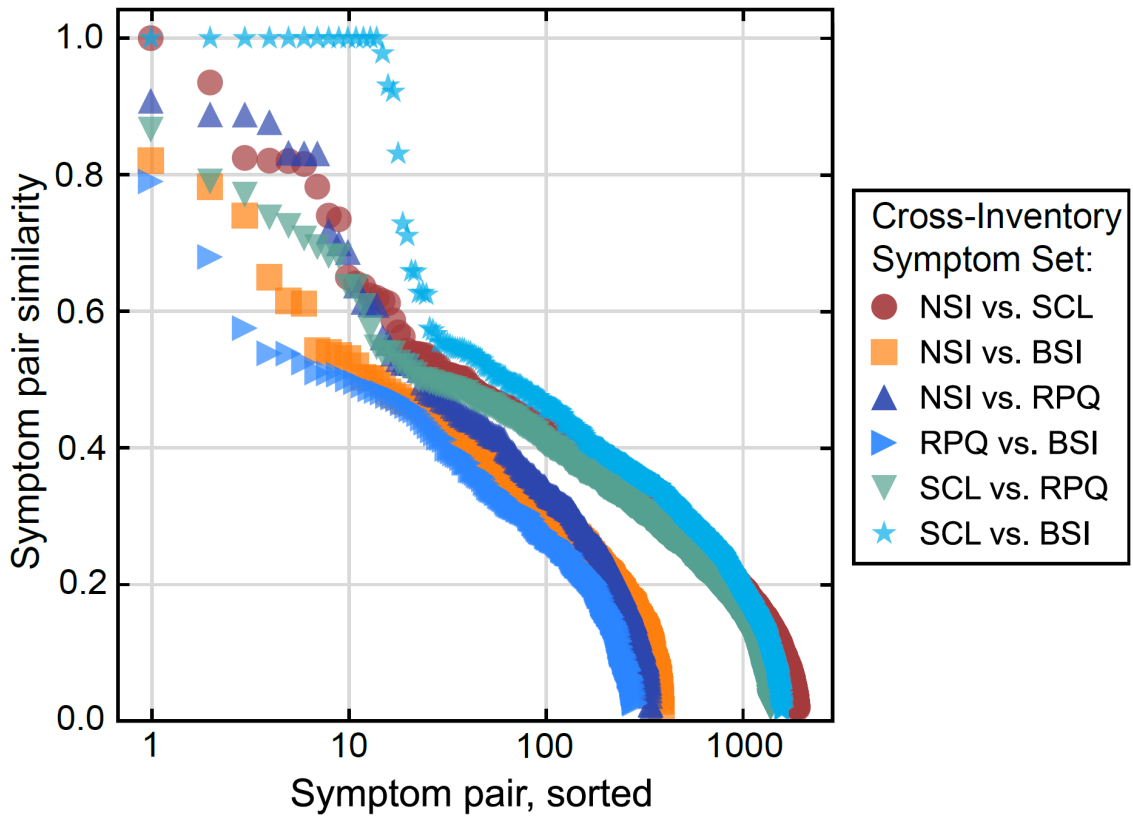

**Figure 4: Quantifying the similarity of symptom inventories.** Boxplots show the distribution of maximum semantic text similarities of symptoms broken out for each inventory crosswalk in both directions. Overall, 5 of the 12 inventory comparisons showed maximum linked text similarity medians of 0.6 or higher. BSI-18: Brief Symptom Inventory–18, RPQ: Rivermead Post Concussion Symptoms Questionnaire, SCL-90-R: Symptom Checklist-90-Revised, NSI: Neurobehavioral Symptom Inventory.

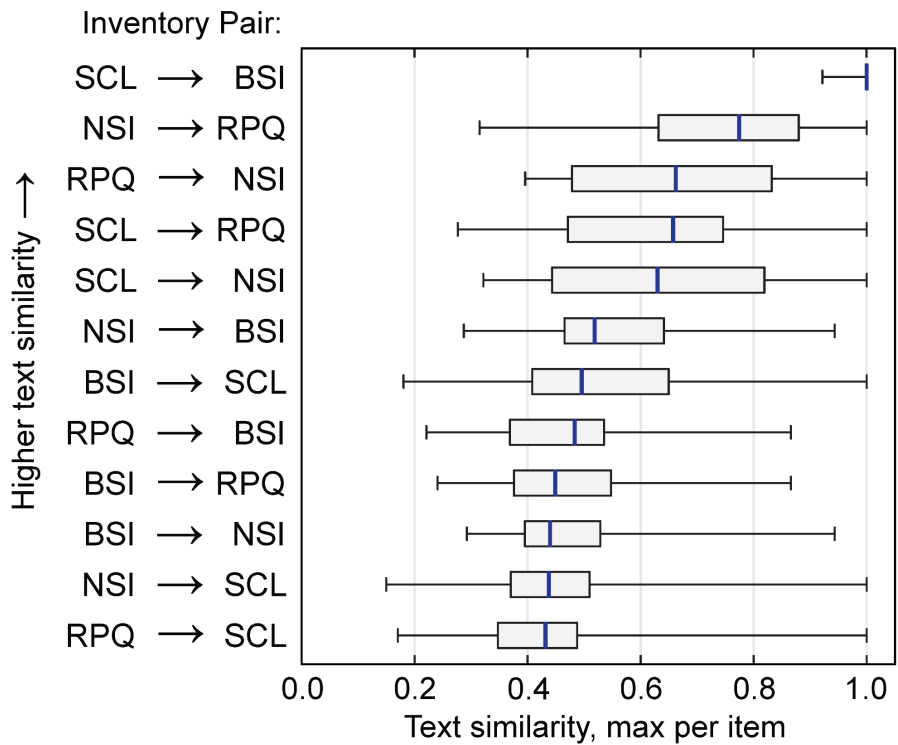

**Figure 5: Cross-inventory symptom prediction.** Cross-inventory accuracies are shown for the Semantic Text Similarity (STS) model, alongside three comparison models (OLS: Ordinary Least Squares regression, RF: Random Forest, GB: Gradient Boosting). Predictions were tested with 50/50 test-train splitting of n=2,056 subjects who were all dually administered both the RPQ and BSI. Only RPQ items were used to estimate BSI item scores (RPQ→BSI, right grey circles), and only BSI items were used to estimate RPQ item scores (BSI→RPQ, right white circles). The non-clinical STS model (blue line) achieved 54.3% EMA, significantly (p<0.001) outperforming other approaches.

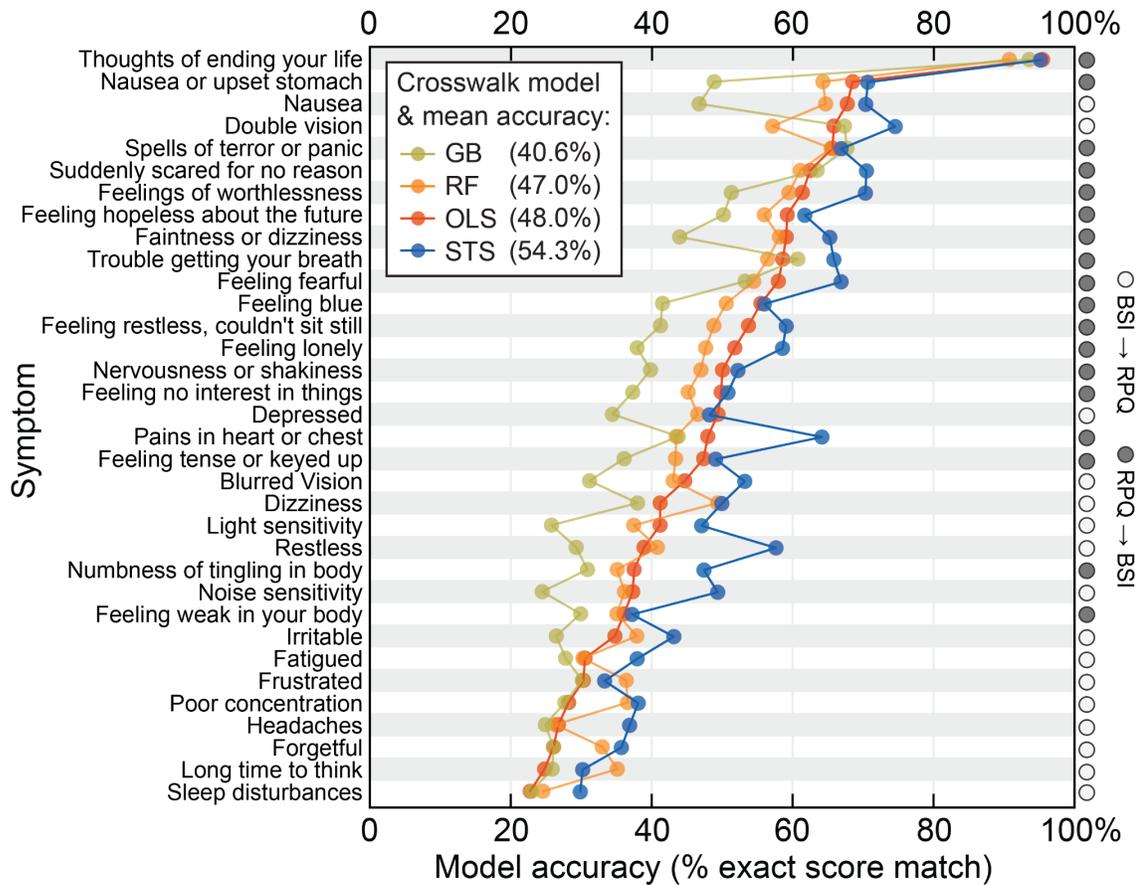

# Supplementary Material for 'Linking Symptom Inventories using Semantic Text Similarity'.

**Contents:**

**Tables**

**Figures**

**Notes**

**Supplementary Table S1: Descriptive characteristics of the data sources.** * indicates data missing.

| Dataset | N | Controls (N) | TBI (N) | F | M | Age range (median) | Symptom Inventory | Missing Variables |
|---|---|---|---|---|---|---|---|---|
| FITBIR-CARE[45–47] | 2188 | 194 | 1994 | 39% | 61% | 17.00 - 25.00 (19.00) | BSI | Education, Race |
| Duke Site#1[48,49] | 34 | 16 | 18 | 24% | 76% | 25.00-57.00 (40.00) | BSI | Ethnicity, Race |
| FITBIR-TRACK TBI[50–53] | 1952 | 238 | 1714 | 33% | 67% | 17.00-88.00 (36.00) | BSI, RPQ | None |
| FITBIR-Okonkwo[54] | 93 | 0 | 93 | 19% | 81% | 22.00-60.00 (34.00) | RPQ | None |
| FITBIR-Lui[55] | 112 | 51 | 61 | 62% | 38% | 18.00-64.00 (31.50) | BSI, RPQ | None |
| LIMBIC-CENC[56,57] | 1529 | 280 | 1246 | 13% | 87% | 22.00-71.00 (38.00) | NSI | None |
| FITBIR-King | 132 | 65 | 67 | 64% | 36% | 18.00-60.00 (37.00) | NSI | Education |
| iSCORE[58] | 100 | 71 | 29 | 13% | 87% | 19.00-51.00 (36.50) | NSI | Ethnicity, Race |
| Longitudinal Chronic TBI in Veterans[59] | 79 | 15 | 64 | 10% | 90% | 23.00-54.00 (32.00) | NSI | Ethnicity, Race |
| Blast-related TBI - Cleveland[60,61] | 44 | * | * | 0% | 100% | 29.00-52.00 (39.50) | NSI | All |
| Blast-related TBI - Houston[60,61] | 35 | * | * | 29% | 71% | 19.00-46.00 (29.00) | NSI | TBI |
| SPIRE | 30 | 7 | 23 | 0% | 100% | 26.00-50.00 (34.50) | NSI | Education, Race, Ethnicity |
| FITBIR-Gill | 18 | 0 | 0 | 39% | 61% | 22.00-60.00 (34.50) | NSI | Ethnicity, Race, TBI |
| fMRI Blast Injury[62] | 17 | * | * | 0% | 100% | 31.00-45.00 (37.00) | NSI | All |
| DoD-ADNI[63,64] | 161 | 71 | 90 | 0% | 100% | 61.00-84.00 (68.00) | SCL | None |
| Duke Site#2[48,49] | 83 | 34 | 49 | 31% | 69% | 23.00-67.00 (36.00) | SCL | Ethnicity, Race |

| Model | t>0 | t>1 | t>2 | t>3 | EMA | Average |
|---|---|---|---|---|---|---|
| Linear Regression | 71.0% | 80.2% | 87.9% | 95.1% | 48.0% | 74.4% |
| Gradient Boosting | 63.2% | 76.4% | 86.6% | 94.9% | 40.6% | 69.4% |
| Random Forest | 70.2% | 78.9% | 87.6% | 95.0% | 47.0% | 73.6% |
| MiniLMBERT | 71.2% | 76.9% | 85.0% | 92.8% | **54.3%** | **74.8%** |
| VAMetadataBERT | 69.2% | 75.0% | 83.4% | 92.5% | 52.7% | 73.3% |
| ClinicalCovidBERT | 70.0% | 75.6% | 84.1% | 92.7% | 53.2% | 73.8% |
| VAClinicalDocsBERT | 69.0% | 74.7% | 83.4% | 92.3% | 52.0% | 72.9% |

**Supplementary Table S2: Cross-inventory model accuracies for seven models.** Three benchmark models (Linear OLS, Gradient boosting, and Random Forest), and four pre-trained variants of STS model were evaluated for accuracy under different scenarios. The models are assessed under five different scenarios: Prediction of binary classification at all four possible threshold scores (0,1,2,3), and EMA. Predictions were tested with 50/50 test-train splitting of n=2,056 subjects who were all dually administered both the RPQ and BSI. The STS model trained on generic, non-medical text (*MiniLMBERT*) performed with the highest accuracy overall (74.8%), although linear regression performed optimally at some thresholds. *MiniLMBERT* also performed optimally on the clinically useful EMA metric, which does not presuppose a specific threshold to assess conversion performance.

**Supplementary Figure S1: Comparing multinomial exact match accuracy (EMA) to mean absolute error (MAE).** Scatterplots show a comparison of two model metrics; the MAE of predicted scores, and multinomial accuracy on a Likert scale. Values were derived for subjects dually administered the RPQ and BSI (n=2,056). Metrics were calculated for each RPQ and BSI item (a), and for each subject (b). EMA is closely anticorrelated with mean absolute error.

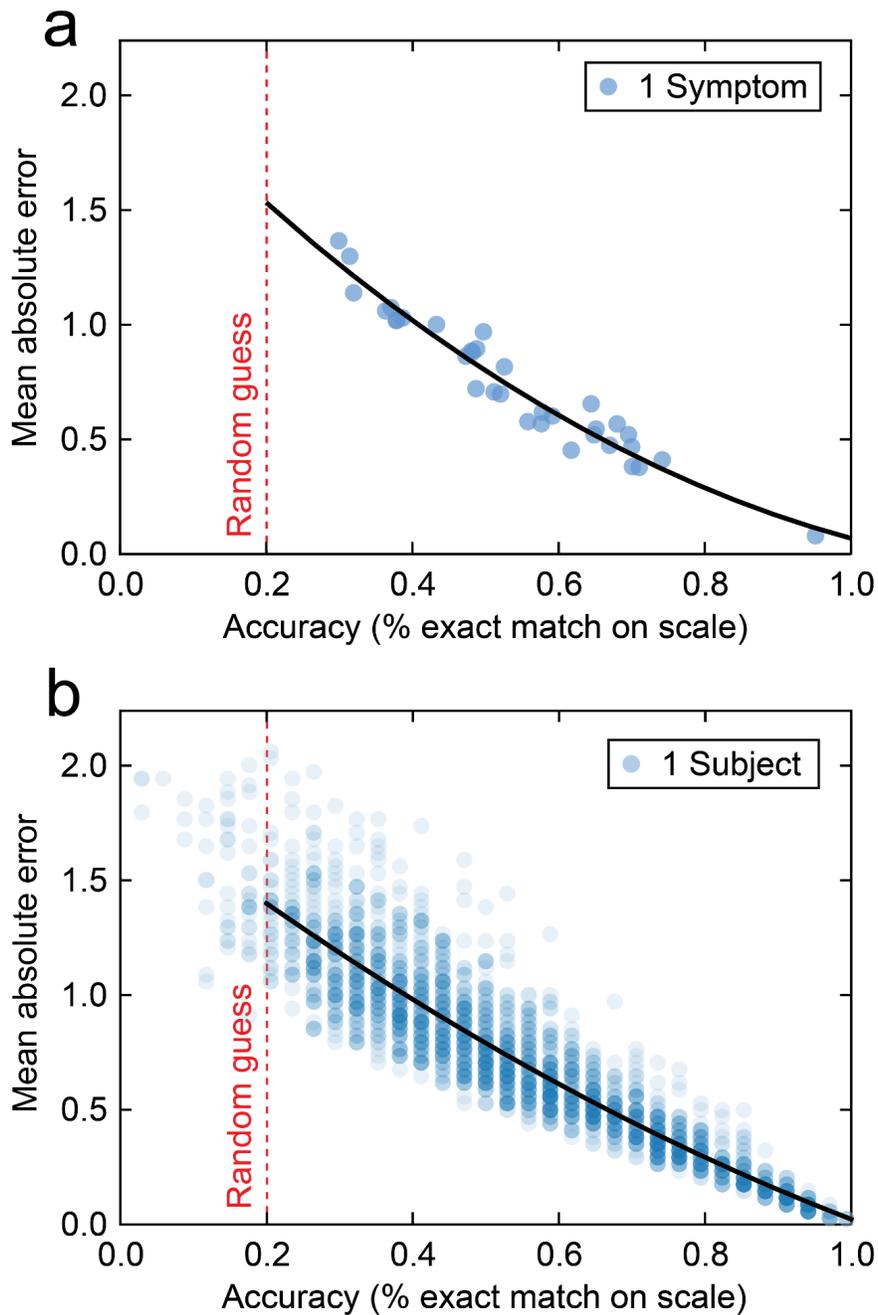

**Supplemental Figure S2: Conceptual illustration of the percentile-based method to adjust for different response scales.** Stacked bar graphs show the percentile distributions of 5-point scale scores observed for different items of inventories. At **(a)** it is shown graphically that an SCL score of 2 is equivalent to an NSI score of either 2 or 3 in percentile terms. Therefore, for SCL→NSI, a score of 2 on the SCL item is converted to a score of either 2 or 3 on the related NSI item. The model selects one of these two values at random, but weighted in proportion to their overlap within a percentile bound (dotted lines). **(b)** Like **(a)**, but for an NSI score of 1, crosswalked NSI→RPQ. An RPQ score of 1 is the most likely model decision, but a converted RPQ score of 0 or 2 is also possible at lower probability (dotted lines). A routine stochastic simulation was used to perform all possible scale crosswalks for all items. This process required only four percentile thresholds per item, and did not involve training.

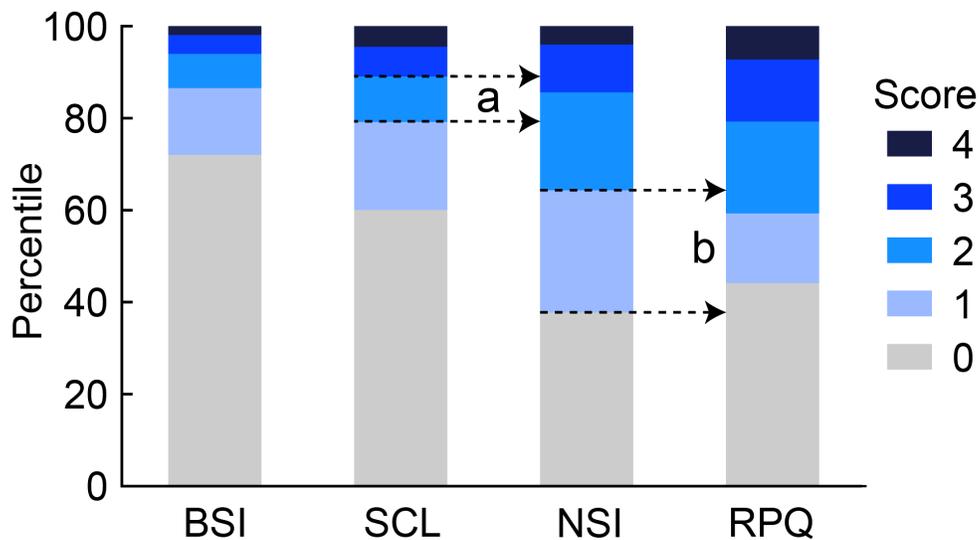

**Supplemental Figure S3: Within-inventory symptom score prediction.** Multinomial accuracies are shown for predicting items within different inventories as a function of randomly sampled numbers of items made available for training. Internal symptom prediction accuracy varied significantly by inventory. EMA improved with the number of items in the inventory that were permitted for training. The average global within-inventory prediction accuracy was 57.7%, compared to the random guess accuracy of 20%.

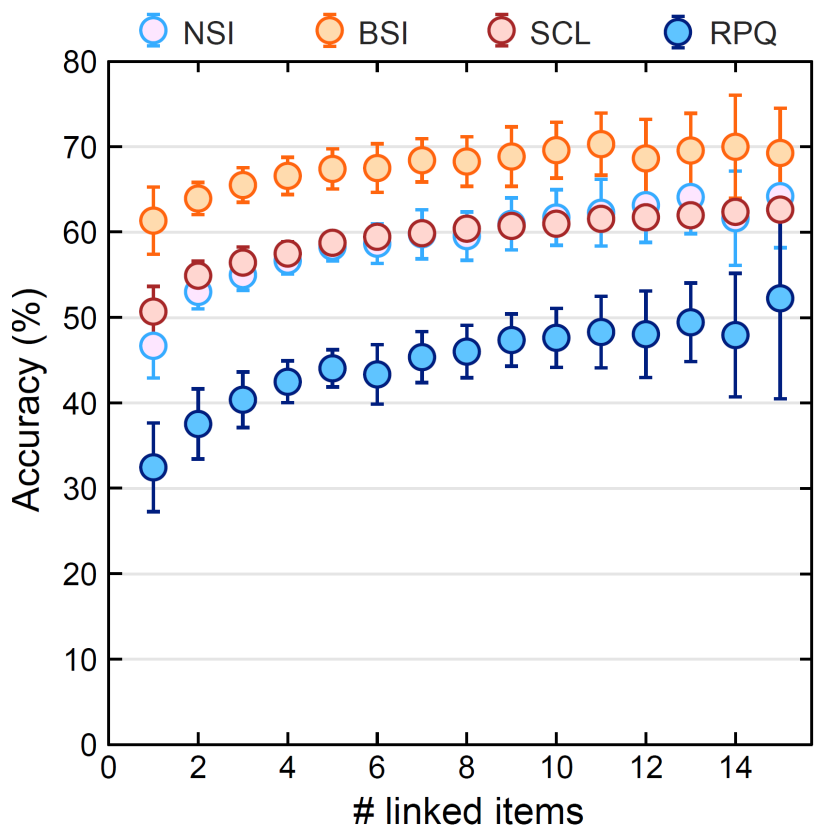

**Supplementary Figure S4:** Annotated inventory conversion website interface.

# Symptom Inventory Crosswalk Interface

## Symptom Inventories Score Conversion

- User selection of recorded inventory → BSI
- User selection of estimated inventory → RPQ
- SHOW
- User input of participant scores →

**BSI Table**
*Enter integer values between 1 and 5*

Anxiety
| Item | Score |
|---|---|
| Feeling tense or keyed up | 1 |
| Nervousness or shakiness | 2 |
| Suddenly scared for no reason | 1 |
| Spells of terror or panic | 3 |
| Feeling so restless you couldn't sit still | 1 |
| Feeling fearful | 3 |

Depression
| Item | Score |
|---|---|
| Feeling no interest in things | 1 |
| Feeling lonely | 4 |

**RPQ Table**
*Output symptoms — Estimated Scores*

Affective
| Symptom | Score |
|---|---|
| Being irritable, easily angered | 3 |
| Feeling depressed or tearful | 5 |
| Feeling frustrated or impatient | 1 |
| Restlessness | 1 |

Cognitive
| Symptom | Score |
|---|---|
| Forgetfulness, poor memory | 3 |
| Poor concentration | 3 |
| Taking longer to think | 1 |

Somatization
| Symptom | Score |
|---|---|
| Light sensitivity, easily upset by bright light | 2 |

- Run conversion → GO
- Converted participant scores ↑

**Supplemental Note 1: Symptom Inventories.**

*Neurobehavioral Symptom Inventory*

The 22-item Neurobehavioral Symptom Inventory (NSI[4]) evaluates cognitive, somatic, and emotional symptoms commonly experienced by adults following a brain injury, including headache, dizziness, memory problems, sleep disturbances, irritability, and difficulty concentrating. Symptom frequency and severity are measured using a five-point Likert-Scale, where the respondent indicates their degree to which they were disturbed by each symptom over the past two weeks from 0 *(None)* to 4 *(Very severe)*.

A total symptom severity score is calculated from the sum of all item responses (range = 0–78). Prior work indicates a large effect size (report ES here) in NSI total scores when distinguishing TBI from non-TBI samples.[65] A number of studies evaluating the factor structure of the NSI in TBI samples found 3-[66,67], 4-[68,69], and 6-factor[70] solutions. Several psychometric studies established that the NSI has excellent internal consistency of total scores in civilian and military mild TBI[71–73] and in adults with chronic mild to severe TBI.[65,67] Further, the NSI possesses excellent test-retest reliability over a 7-day period[72] and adequate test-retest reliability over a 30-day period[74] in mild TBI.

*Rivermead Post-Concussion Symptoms Questionnaire*

The 16-item Rivermead Post-Concussion Symptoms Questionnaire (RPQ[2,3]) measures the presence and severity of commonly reported TBI symptoms, including headache, dizziness, fatigue, irritability, and concentration difficulties. Using a five-point Likert-Scale from 0 (*not experienced*) to 4 (*severe problem*), respondents are instructed to rate the severity of each symptom experienced within the last 24 hours, relative to their experience of the symptom before injury. The total score for the RPQ (range = 0–64) is determined by summing the scores across all symptoms that are rated at a level of mild severity or above (i.e., with a score of 2 or higher). The RPQ can predict moderate to severe limitations in psychosocial adjustment and participation in daily activities in acute[75] and chronic[76] mild TBI. The RPQ has well-established internal consistency,[77] test-retest reliability,[2] and internal construct validity.[77,78]

*Brief Symptom Inventory–18*

The 18-item Brief Symptom Inventory - 18 Item version (BSI-18[5]) is a shortened version of the Symptom Checklist-90-Revised (SCL-90-R) and the original 53-item BSI designed to efficiently assess psychological symptoms and distress in both healthy and patient populations. The BSI-18 consists of items rated on a five-point scale, ranging from 0 (*not at all*) to 4 (*extremely*), indicating how much the problems distressed or bothered respondents over the past seven days.[5] Scoring the BSI-18 involves summing the item responses within three primary symptom dimensions; somatization, depression, and anxiety. Higher scores on each subscale indicate a higher level of symptom severity, and an overall Global Severity Index (GSI) score can be calculated by summing all item responses. The BSI-18 has demonstrated good reliability, with high internal consistency reported for the three dimensions,[79,80]. Test-retest reliability has also proven satisfactory over short time intervals.[80] The BSI-18 has shown strong convergent validity, with significant correlations between its dimensions and subscales of the SCL-90-R.[81]

*Symptom Checklist-90-Revised*

The Symptom Checklist-90-Revised (SCL-90-R[6]) is a widely used self-report measure designed to assess the presence, severity, and frequency of a broad range of psychological symptoms and emotional distress, and like the BSI-18, is not just specific to TBI. Respondents rate each item on a five-point scale from 0 (*Not at all)* to 4 (*extremely*), indicating how much they have been bothered or distressed by the symptom over the past week. The responses to individual items are summed to obtain scores for nine primary symptom dimensions, and a global average measure can be also calculated as the total score / 90. The SCL-90-R demonstrates good internal consistency for the different symptom dimensions,[81,82] and measured test-retest reliability is satisfactory.[83] The SCL-90-R exhibits strong convergent validity, and good correlation between its dimensions and other established measures of psychopathology.[79,81] The SCL-90-R has been widely validated across various patient populations and cultural contexts.

**Supplemental Note 2: Scoring unrelated phrases containing the same words.**

This study relies on the ability of transformer models to measure true semantic similarity, as distinct from overlapping/similar words across text descriptions. To assess this distinction, we tested sentence pairs with overlapping words but different meaning, and sentence pairs with similar meaning but no overlapping words. For example, using the primary sentence "*Medspacy is a powerful library of clinical language processing tools*", we tested against two sentences 1 (related). "*It is a repository of methods for analysing medical text*" and 2 (unrelated). "*The ability to process language is powerful when you are in the library*". The unrelated sentence contains four of the same words as the primary sentence "*powerful*", "*library*", "*process*", "*language*", whereas the related sentence contains no overlapping words. *MiniLMBERT* correctly scored sentence 1 as S = 0.59 against the primary sentence, and sentence 2 as S = 0.32 against the primary sentence. Across many examples, transformer models were consistently sensitive to the semantic similarity of sentence pairs, and were insensitive to overlapping words across sentences with semantically different use and meanings.